\documentstyle[12pt,psfig,named]{article}

\title{TAKTAG: Two-phase learning method for hybrid statistical/rule-based
part-of-speech disambiguation}
\author{Geunbae Lee\\ Jong-Hyeok Lee \\ Sanghyun Shin \\
Department of Computer Science \& Engineering \\
and Postech Information Research Laboratory \\
Pohang University of Science \& Technology \\
San 31, Hoja-Dong, Pohang, 790-784, Korea \\
gblee@vision.postech.ac.kr
}

\date{}

\begin{document}

\maketitle

\begin{abstract}
Both statistical and rule-based approaches to part-of-speech (POS)
disambiguation have their own advantages and limitations. Especially for
Korean, the narrow windows provided by hidden markov model (HMM) cannot cover
the necessary lexical and long-distance dependencies for POS disambiguation.
On the other hand, the
rule-based approaches are not accurate and flexible to new tag-sets and
languages. In this regard, the statistical/rule-based hybrid method
that can take
advantages of both approaches is called for the robust and flexible POS
disambiguation. We present one of such method, that is, a two-phase
learning architecture for the hybrid statistical/rule-based
POS disambiguation, especially for Korean.
In this method, the statistical learning of morphological tagging is
error-corrected by
the rule-based learning of Brill [1992] style tagger. We also design the
hierarchical and flexible Korean tag-set to cope with the multiple
tagging applications, each of which requires different tag-set.
Our experiments show
that the two-phase learning method can overcome the undesirable features
of solely HMM-based or solely rule-based tagging, especially for
morphologically
complex Korean.
\end{abstract}

\section{Introduction}
Part-of-speech (POS) tagging is a basic step to several natural language
processing applications including text-based information retrieval, speech
recognition, and text-to-speech synthesis. The POS tagging has been
usually performed by statistical (or data/corpus-driven) approaches mainly
using hidden markov model (HMM)
\cite{church:stochastic,cutting:practical,kupiec:robust,weischedel:coping}.
However, since statistical approaches only consider the neighboring tags
within a limited window (usually two or three),
sometimes the decision cannot cover all the linguistic rules necessary for the
disambiguation. Also the approaches are inappropriate for the idiomatic
expressions in which the lexical term itself needs to be consulted for the
disambiguation. The statistical approaches are insufficient for the
agglutinative languages (such as Korean) which have usually complex
morphological structures.
In these languages, a word consists of single stem morpheme plus several
functional
morphemes, and the POS tags should be assigned to each morpheme to best
exploit the complex morphological structures.
Considering just the neighboring morphemes regardless of their grammatical
functions is not enough for the morpheme-level POS disambiguation.
Recently, rule-based approaches are re-studied to cope with the limitations
of statistical approaches by learning the tagging rules automatically from
the corpus \cite{brill:simple,brill:some}. Some systems even perform the POS
tagging as part of syntactic analysis process \cite{voutilainen:syntax}.
However, the rule-based
approaches alone are in general not robust to handle the unknown words,
and is not flexible to adjust to the new tag-sets and
languages. Also the performance is usually no better than the statistical
counterparts \cite{brill:simple}. To gain flexibility and robustness
and also to overcome the
limited window range of statistical approaches, we need a method that can
combine both statistical and rule-based approaches \cite{tapanainen:tag}.

This paper presents a hybrid POS disambiguation methods that cascaded
statistical and
rule-based approaches in a two-phase learning architecture.
Our system TAKTAG (Two-phase learning Architecture for Korean part-of-speech
TAGger) combines the state-of-the-art hidden markov model with
Brill [1992] style rule learning error correction. The system is trained in
two phases: HMM parameter estimation and comparison-based rule
learning for the HMM tagging output. The TAKTAG has the unique following
properties of the Korean POS disambiguation:
\begin{itemize}
\item The system is designed to be very accurate in tagging especially the
ambiguous
Korean morphemes that have more than one part-of-speeches. The accuracy is
very important in Korean tagging since Korean has much poorer tagging
performance compared with English due to its linguistic characteristics.
Although some of
the POS ambiguities cannot be resolved at the morphology level, we
tried to correct as much as tagging errors by introducing the rule-based
error correction scheme.
\item The system fully considers many linguistic characteristics of
Korean in HMM/rule tagging. Unlike English and other Indo-European languages,
the
complex functional morphemes determine the grammatical roles of Korean
words (which is called Eojeol, see section~\ref{sec:hier}).
\item The system is flexible so that it can tune to the new tag-sets and
new languages. In other words, the system doesn't rely on the enormous
amounts of pre-existing
tagged corpus for its training. This is very important since the Korean tag
sets are not stabilized yet, nor are the standard Korean tagged corpus
provided yet.
\item In TAKTAG, the tag-sets are hierarchically organized so that they can
be adjustable according to the given applications such as
information retrieval, speech synthesis, text data extraction, and so on.
\end{itemize}
The rest of the sections are organized as follows.
Section~\ref{sec:hier} explains the linguistic characteristics of Korean and
the
hierarchically organized tag sets for multiple applications.
Section~\ref{sec:two} discusses the two-phase learning architecture, its
process model and the training procedures. Section~\ref{sec:exp} demonstrates
the performance of TAKTAG with extensive experiments and finally
section~\ref{sec:con} draws some conclusions of the works.

\section{Hierarchical tag-sets for Korean morphology}
\label{sec:hier}
Korean is classified as agglutinative languages in which the words (which is
called Eojeol in Korean) consist
of several morphemes that have clear-cut morpheme boundaries. Below are the
characteristic of Korean that must be considered for POS tag-set and tagging
system design.
\begin{enumerate}
\item Korean is a postpositional language
with many kinds of noun-endings, verb-endings, and
prefinal verb-endings. It is the functional morphemes, not Eojeol's order that
determine the most of the grammatical relations such as the noun's
case roles, verb's tenses, modals, and modification relations between
Eojeols. So contextual information for POS disambiguation must be
selectively applied to the functional (bound) morphemes or content (free)
morphemes.
\item Sometimes a Korean Eojoel corresponds to an English phrase, not to a
single
word, so the tagging must be done on morpheme basis, not Eojeol basis. The
morphological analyzer must precede the tagging system because the
morpheme segmentation is much more important and difficult than POS
assignment in Korean.
\item Korean is basically SOV
languages but has relatively free word order
compared to English, except for the constraints that the verb must appear
in a sentence-final position. However, in Korean, some word-order
constraints do exist
such that the modifiers must be placed before the word (called head) they
modify. So some order constraints must be applied as contextual
information, but some must not.
\item Complex spelling changes can occur between morphemes when two morphemes
combine to form an Eojeol. These spelling changes make it difficult to
segment the morphemes before assigning the POS tags. Also a lot of allomorphs
are generated from the spelling changes.
\end{enumerate}
For the above reasons, a morphological analysis play important roles in Korean
POS tagging system. It is the morphological analysis process which
initially segments
the morphemes out of the Eojoels, reconstructs the spelling changes, and
assigns the initial POS tags to each morpheme by consulting the dictionary.
Later, the tagging system disambiguates the POS assignments by selecting
the single morpheme sequence for each sentence and the single POS tag
for each morpheme by
consulting the lexical and contextual information acquired from the corpus.

We classified over 200 POS tags that can be used in
morphological analysis as
well as the POS disambiguation. Our POS tags, which are originally
designed for
morphotactics modeling in CYK-based Korean morphological
analysis \cite{lee:the}, consists of the
hierarchically organized 200 symbols that are refined from the seven major
grammatical categories of Korean, which are nominal, predicate, modifier,
particle, ending, symbol, interjection. For single morpheme,
a path name in the POS symbol hierarchy
(e.g. nominal:noun:proper-noun:person-name:no-final-consonant)
is assigned as a POS tag.
The tag can be a full path name or part of the path name to adjust the
number of tags in the tag-set.
In this way, the tag-set can be adjusted by refining the
more pertinent grammatical categories to the applications at hand. For
example, for the text information retrieval application, we can more refine the
nominals than the predicates since the indexing terms are usually nominals.
Figure~\ref{tb:tagset} shows one example of tag-set extracted from the POS
symbol hierarchy. This tag-set will be used in our experiment in
section~\ref{sec:exp}.

\begin{table}
\begin{center}
\begin{tabular}{|c|c|c|c|} \hline
tag & description & tag & description \\ \hline
MP & proper noun & jC & case particle \\
MD & bound noun & jJ & conjunctive particle \\
MC & common noun & jS & auxiliary particle \\
S & numeral & mC & conjunctive ending \\
T & pronoun & mT & final ending \\
D & verb & mj & derivative ending \\
H & adjective & - & suffix \\
G & adnoun & + & prefix \\
B & adverb & e & prefinal ending \\
y & predicate particle & & \\ \hline
\end{tabular}
\end{center}
\caption{Example tag-set derived from the hierarchical part-of-speech
symbols for Korean.}
\label{tb:tagset}
\end{table}

\section{Two-phase learning of POS disambiguation}
\label{sec:two}
\begin{figure}
\centerline{\psfig{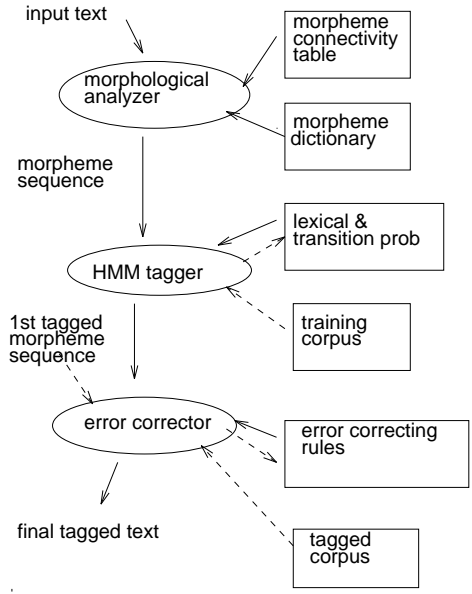}}
\caption{Two-phase learning architecture for Korean POS tagging. The ovals
designate the process, and the boxes are resources for the process. The
solid arrows are for performance phase, but the dotted arrows are for
training phase. The first-tagged-morpheme-sequence's are the output of the
HMM tagger as well as part of the training corpus for the error-corrector.}
\label{fg:arch}
\end{figure}

Figure~\ref{fg:arch} shows a two-phase learning architecture for Korean
POS tagging
system. There are three major components: the morphological analyzer,
the HMM tagger, and the error-corrector.
The morphological analyzer segments the constitutional morphemes out
of the input
texts and assigns the initial POS tag for each morpheme by
consulting the dictionary.
The Korean morphological analysis procedure consists of the following
three steps:
morpheme segmentation, morphotactics modeling, spelling change handling
\cite{sproat:morph}. The input texts are scanned from left to right,
character by character, to be matched to the morphemes in the dictionary. For
the efficient text search, the modified CYK parsing method based on the
dynamic programming technique is applied \cite{lee:the}. For the
morphotactics modeling, we defined the hierarchical POS symbol system as
mentioned in section~\ref{sec:hier}.
At the bottom of the hierarchy, there are about 200 POS tags that
reflects the morphosyntactic properties of Korean. The full path name in
the POS symbol hierarchy is
encoded in the dictionary for each morpheme entry, and is called morpheme
connectivity information. To model the morpheme's
connectability to each other (so called morphotactics),
the separate morpheme connectivity table encodes
the connectability of each 200 morpheme connectivity information.
So when the input Eojeol is
segmented by the dictionary search, the analyzer checks
whether the segmentation is legal or not by consulting the morpheme
connectivity table to find out the connectability of the two segmented
morphemes.
In the dictionary, we also enroll the inflected forms as
well as original (uninflected) morphemes as header information, so that we
can reconstruct the original form from the spelling changes of morphological
combination.

The HMM tagger takes the morpheme sequence with the initial tag assignment
by the morphological analyzer, and
using the Viterbi algorithm \cite{forney:viterbi}, searches the optimal tag
sequence for the POS disambiguation. Sometimes, there can be multiple morpheme
sequences for one sentence due to the multiple segmentation results in
Korean.
In that case, we perform Viterbi search for each morpheme sequence, and
select the maximum probability tag sequence as a solution.
To reduce the computational complexity, we can share common morphemes in the
different morpheme sequence during the Viterbi search as studied in
\cite{kim:efficient}. The equation of the HMM tagging model we use is the
ordinary bi-gram model with left to right search:
$T^{*} = argmax_{T} \prod_{i=1}^{n} Pr(t_{i}|t_{i-1})Pr(m_{i}|t_{i})$,
where $T^{*}$ is the optimal tag sequence that maximizes the forward
Viterbi scores.
The $Pr(t_{i}|t_{i-1})$ is a bi-gram tag transition probability,
and the $Pr(m_{i}|t_{i})$ is a morpheme lexical probability.

We call the results of this HMM tagging as the
{\em first-tagged-morpheme-sequence},
and usually the tagging accuracy is not satisfactory because of the
characteristics of Korean as mentioned in section~\ref{sec:hier}.
The error-corrector transforms the first tagged morpheme
sequences to the final tagged text. The error-corrector is a rule-based
transformer, and it matches the condition
part of the rules, and change the erroneous tags to the tags in the action
part. The rules are in the form of:
$[current-morpheme][current-tag];
([context-morpheme]:[context-morpheme-or-lexical
form])^{\star} \rightarrow [current-morpheme][corrected-tag]$,
where the rule condition part consists of the current and context morphemes
with their
tags, and the action part is the current morpheme with the corrected tag.
The $\star$ means
that the rule can see the several composite contexts at one time.
The next section
explains the training algorithms of the HMM tagger and the error-corrector in
detail, and shows what kinds of error correcting rules are learned to overcome
the statistical tagging limitations.

\subsection{Learning HMM-based disambiguation}
The first phase of learning in the two-phase POS disambiguation is
the HMM parameter
training. Since the HMM tagger takes morpheme sequences as input, unlike
English, the
training corpus must be morphologically analyzed, too. The POS tags are
assigned to each original morpheme (in the training corpus) which is
reconstructed from the spelling
changes. There can be many morphological analysis results
for one sentence in Korean. In that case, we include only correct
morphological analysis results in the training corpus by following
\cite{kim:efficient}.

There are two types of HMM parameter training methods that are widely used.
The first method is to use the enormous amounts of tagged corpus such as
Brown corpus \cite{francis:frequency} to extract the lexical and
transition probabilities from the
frequences of tags associated with words and of pairs of tag
\cite{church:stochastic}. This method is not desirable at the moment for
Korean because 1) there is no large tagged corpus available yet, and
2) the tag-sets
are not standardized yet. The second method of training does not require
large tagged corpus for training \cite{cutting:practical}. In this case, the
Baum-Welch algorithm \cite{baum:inequality} can be used for estimation of
the HMM parameters by iterative relaxation (which is one form of the
estimation-maximization (EM) algorithm). However, several studies show that
using as much as tagged corpus for training gives much better performance
on tagging
\cite{merialdo:tagging}, and the fact favors for the Church [1988] style
tagging as long as large tagged corpus is available for Korean.
However, for the parameter estimation from the small amount of tagged
corpus, the Baum-Welch algorithm always helps to increase the tagging
performance. In this regard, we used
small amounts of tagged corpus (about 2000 morphemes) for bootstrapping the
Baum-Welch training, and mainly use the Baum-Welch algorithm for the
whole training using the morphologically analyzed but untagged corpus.

\subsection{Learing rule-based tagging correction}
The statistical morpheme tagging only covers the limited range of
contextual information. Moreover, it does not see the lexical form itself in
disambiguation. As mentioned before, Korean has very complex morphological
structure so it is necessary to see the functional morphemes selectively to
get the relation between Eojeols. For these reasons, we designed the error
correcting rules to compensate the missings of the statistical tagging.
However, designing the tagging rules with knowledge engineering is
tedious and error-prone. Instead, we adopted Brill`s approach [1992] to
automatically learn the error correcting rules from the tagged corpus.
Fortunately, Brill showed that we don't need large tagged corpora
to extract the symbolic rules, especially
compared with the ones in need for the statistical tagging. Table~\ref{tb:rule}
shows the rule schema we used to extract the error correcting rules, where
the rule schema designates the context, i.e., the place and the lexical/tag
decision in the rule
(see rule format in section~\ref{sec:two}). More than one rule schema can be
simultaneously applied to the error correction so that the rule can see
more than one contexts at one time.
\begin{table}
\begin{center}
\begin{tabular}{|l|l|} \hline
rule schema & description \\ \hline
N1FMT & next single Eojeol first morpheme's tag \\
P1LMT & previous single Eojeol last morpheme's tag \\
N2FMT & next second Eojeol first morpheme's tag \\
N3FMT & next third Eojeol first morpheme's tag \\
PlLMO & previous single Eojoel last morpheme's lexical form \\
P1FMO & previous single Eojeol first morpheme's lexical form \\
N1FMO & next single Eojeol first morpheme's lexical form \\ \hline
\end{tabular}
\end{center}
\caption{The example rule schema to extract the error correcting rules. The
TAKTAG has about 24 rules schema in this form.}
\label{tb:rule}
\end{table}
The rules are learned according to the schema by comparing the correctly
tagged corpus (morphologically analyzed and hand tagged) with the output
of the HMM tagger (called the first-tagged-morpheme-sequence).
The acquired rules are sorted by their effectiveness
which is defined by the number of successful corrections using the rules
as used in \cite{brill:simple}.

\section{Experiments}
\label{sec:exp}
We collected 70000 morpheme corpus which was from diverse domains such as
national ethics code,
elementary school textbooks, composition handbooks, and so
on\footnote{The 300000 Eojeol (about 700000 morpheme)
untagged corpus was provided from the ETRI (Electronics and Telecommunication
Research Institute) in Korea. We selected 70000 morpheme
sentences among them
with careful consideration to the corpus balance.}. All the sentences
in the corpus
were morphologically analyzed before use. In each domain, about 15\% of the
corpus are manually
tagged for error correcting rule learning, about 15\% are set aside for
the test, and the remaining 70\% are used for the Baum-Welch training. For
initial bootstrapping of HMM, we are provided with other 2000 morpheme tagged
corpus which is disjoint from our original corpus.
{}From the 15\% (about 10000 morphemes) of the corpus, we extracted about
445 error-correction rules using the rule schema in table~\ref{tb:rule}.
Table~\ref{tb:result} shows the final tagging results. The accuracy is
calculated from the formula: $\frac{(number-of-tagged-morphemes) -
(number-of-incorrectly-tagged-morphemes)}{(number-of-tagged-morphemes)}$.
The results show that the error correcting rules are quite useful to
increase the overall tagging accuracy, and the overall results are
much better than the previous well-engineered HMM tagging results (which was
about 89.1\% in a similar environment) even
though our HMM tagging alone are not quite successful \cite{kim:efficient}.
This results demonstrate that the well-engineered HMM tagging with our error
correcting rules can increase the
overall tagging performance up to over 97\% which was considered to be
impossible with statistical tagging alone in English \cite{tapanainen:tag}.
\begin{table}
\begin{center}
\begin{tabular}{|l|l|l|l|l|} \hline
corpus & no. morph. & no. ambig. morph. & HMM alone & two-phase \\ \hline
national ethics & 269 & 151 & 81.8 & 91 \\
composition handbook & 2227 & 710 & 79.7 & 92.5 \\
science text1 & 4660 & 1539 & 78.3 & 91.2 \\
science text2 & 3973 & 1329 & 78.1 & 93 \\ \hline
total & 11129 & 3729 & 79.5 & 91.9 \\ \hline
\end{tabular}
\end{center}
\caption{The experiment results. From the left, corpus category, number of
morphemes in the corpus, number of ambiguous morphemes that have more than
one POS, HMM tagging performance (\%), two-phase learned tagging performance
(\%).}
\label{tb:result}
\end{table}

\section{Conclusions}
\label{sec:con}
We presented a new POS tagging architecture which
integrates the statistical approach with the rule learning approach in a
synergistic way. Our hybrid tagging architecture is proved to be useful,
especially for the morphologically complex agglutinative languages such as
Korean. The system TAKTAG can provide the
following two unique properties for desirable Korean tagging: 1) The system can
provide accurate results even with the morpheme
tagging which usually results in very poor performance, and 2) The system can
be flexibly tuned to the new tag-sets without massive retraining.
The performance of the two-phase learning for tagging is determined how well
the error-corrector can compensate the deficiencies of the statistical
tagging, and in that sense, our TAKTAG is much successful since it increased
the overall tagging results more than 10\%. The next step will be to analyze
the learned rules carefully to extract the more desirable rule schema for
Korean.
The robust unknown word handling scheme with more efficient morphological
analyzers also should be studied with the well-engineered HMM taggers that
fully consider the linguistic characteristics of Korean.

\section*{Acknowledgments}
This research was partially supported by POSCO (Pohang iron and steel
company). We thank to NamHee Hong and Wonil Lee for re-classification of Korean
part-of-speech and re-implementation of Korean morphological analyzer.
The corpus was selected from the one provided by ETRI (Electronic and
Telecommunication Research Institute), Korea.

\bibliography{/mnt/home/1/HCI/gblee/paper/nlpsp}
\bibliographystyle{named}
\end{document}